\documentclass[lettersize,journal]{IEEEtran}
\usepackage{amsmath,amsfonts}
\usepackage{algorithmic}
\usepackage{algorithm}
\usepackage{array}
\usepackage[caption=false,font=normalsize,labelfont=sf,textfont=sf]{subfig}
\usepackage{textcomp}
\usepackage{stfloats}
\usepackage{url}
\usepackage{verbatim}
\usepackage{graphicx}
\usepackage{cite}
\usepackage{amssymb} 
\usepackage{booktabs}
\usepackage{multirow}
\usepackage{xcolor}
\begin{document}

\title{ZSG-IAD: A Multimodal Framework for Zero-Shot Grounded Industrial Anomaly Detection}

\author{Qiuhui Chen, Jiaxiang Song, Shuai Tan, and Weimin Zhong%
\thanks{This work was supported in part by the Fundamental and Interdisciplinary Disciplines Breakthrough Plan of the Ministry of Education of China under Grant JYB2025XDXM402 and in part by the National Natural Science Foundation of China under Grant U25A20468. (Corresponding author: Weimin Zhong.)}%
\thanks{Qiuhui Chen, Jiaxiang Song, Shuai Tan, and Weimin Zhong are with the School of Information Science and Engineering, East China University of Science and Technology, Shanghai, China (e-mail: chenqh@ecust.edu.cn; songjiaxiang@ecust.edu.cn; tanshuai@ecust.edu.cn; wmzhong@ecust.edu.cn).}%
}



\maketitle

\begin{abstract}
Deep learning–based industrial anomaly detectors often behave as black boxes, making it hard to justify decisions with physically meaningful defect evidence. We propose ZSG-IAD, a multimodal vision–language framework for zero-shot grounded industrial anomaly detection. Given RGB images, sensor images, and 3D point clouds, ZSG-IAD generates structured anomaly reports and pixel-level anomaly masks.
ZSG-IAD introduces a language-guided two-hop grounding module: (1) anomaly-related sentences select evidence-like latent slots distilled from multimodal features, yielding coarse spatial support; (2) selected slots modulate feature maps via channel–spatial gating and a lightweight decoder to produce fine-grained masks. To improve reliability, we further apply Executable-Rule GRPO with verifiable rewards to promote structured outputs, anomaly–region consistency, and reasoning–conclusion coherence. Experiments across multiple industrial anomaly benchmarks show strong zero-shot performance and more transparent, physically grounded explanations than prior methods. We will release code and annotations to support future research on trustworthy industrial anomaly detection systems.
\end{abstract}


\section{Introduction}

\begin{figure}[tbp]
    \centering
    \includegraphics[width=\linewidth]{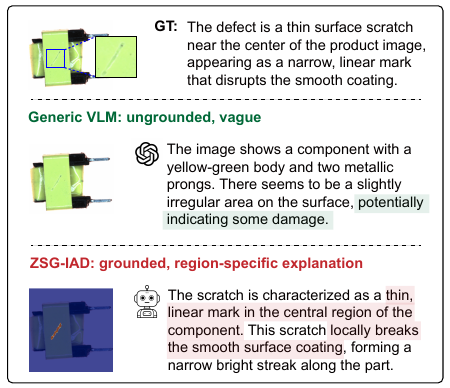}
    \caption{From vague anomaly descriptions (generic VLM, GPT-4o~\cite{hurst2024gpt}) to grounded, region-specific reports (ZSG-IAD) on a scratched industrial part.}
    \label{fig:abstract}
\end{figure}

Industrial anomaly detection (IAD) is a core capability in smart manufacturing, where defective products and faulty components must be identified quickly and reliably to ensure quality control and operational safety.
In practical factories, inspection decisions are rarely made from a single modality: engineers routinely cross-check RGB visual images, sensor-derived images (e.g., infrared or pseudo-3D cues), and 3D measurements (e.g., point clouds) to determine \emph{what} defect is present, \emph{where} it is located, and \emph{why} the conclusion is justified.
However, most existing IAD systems still behave as black boxes, typically producing only an anomaly score or a coarse heatmap.
Such outputs are often insufficient for industrial auditing, root-cause analysis, and closed-loop rework, where decisions must be traceable to physically meaningful evidence and defect regions.

Recent progress in vision-language models (VLMs) and large language models (LLMs)~\cite{Radford_2021_ICML,openai2023gpt4} suggests a promising direction: generating natural-language explanations for inspection results.
Nevertheless, directly applying generic multimodal LLMs to industrial inspection remains problematic.
First, industrial anomalies are usually \emph{non-semantic} and fine-grained (e.g., subtle scratches, contamination, micro-cracks), making it hard for general models to produce precise, evidence-based descriptions.
Second, descriptions produced by prompting often become vague and ungrounded: the text may not explicitly align with defect regions, and the stated evidence may not match the model's localization behavior (Fig.~\ref{fig:abstract}).
Third, real deployment typically requires standardized report formats (e.g., defect type/location/confidence) for downstream QA workflows, yet free-form generation lacks enforceable protocol constraints, leading to inconsistent and hard-to-verify outputs.

To address these challenges, we propose \textbf{ZSG-IAD}, a multimodal vision--language framework for \emph{zero-shot grounded} industrial anomaly.
Here, “zero-shot” refers to dataset-level transfer: the model is trained on one set of object categories and evaluated on disjoint categories, without any fine-tuning on the target benchmark.
Given RGB images, sensor images, and 3D point clouds, ZSG-IAD outputs:
(i) a machine-checkable \emph{structured anomaly report} and
(ii) a \emph{pixel-level anomaly mask} for localization.
ZSG-IAD follows a ``frozen-encoder + lightweight-adapter'' design for efficient adaptation: frozen 2D/3D encoders extract robust representations, while lightweight modules are trained to enhance cross-modal interaction and align features to a causal LLM for structured reporting.
In particular, we strengthen 2D cross-source interaction between RGB and sensor features via Channel-Spatial Swapping (CSS), and align 2D-3D representations through a bidirectional cross-attention fusion layer.
To support region-aware reporting, we distill a compact set of evidence-like latent tokens using Slot Attention, which are injected into a LoRA-adapted causal LLM to generate structured outputs.

We introduce a \textbf{language-guided two-hop detection-to-region grounding} head that maps an anomaly-related sentence to a defect region.
In Hop-1 (Text$\rightarrow$Slots), the sentence representation softly selects evidence-like slot features and aggregates their coarse spatial support map.
In Hop-2 (Slots$\rightarrow$Pixels), the selected slot feature and coarse map jointly modulate 2D feature maps via channel--spatial gating and a lightweight decoder, producing a fine-grained pixel-level anomaly mask.
This explicit coarse-to-fine design improves interpretability: the intermediate coarse map exposes where the model ``looks'' when forming a defect claim, while the refined mask provides actionable localization for inspection and rework.

Beyond supervised learning, we further improve protocol adherence and diagnostic faithfulness using reinforcement fine-tuning.
Specifically, we adopt \textbf{Executable-Rule GRPO} under an RL-with-Verifiable-Rewards (RLVR) paradigm:
during GRPO, only LoRA parameters are updated while a frozen SFT model serves as the reference policy.
We define fully automatic, executable rewards that jointly enforce:
(i) structured reporting validity, (ii) grounding alignment between the Hop-1 coarse map and the final mask, and
(iii) logical consistency between \texttt{<Reasoning>} and \texttt{<DefectType>}.
This reinforcement stage reduces format violations and mitigates mismatches between textual claims and localization behavior, yielding more reliable and auditable reports without requiring human preference labels.
Our main contributions are:
\begin{itemize}
    \item \textbf{A multimodal framework for zero-shot grounded IAD.}
    We propose ZSG-IAD, an end-to-end vision--language system that consumes RGB, sensor images, and 3D point clouds, and produces both structured anomaly reports and pixel-level defect masks for transparent industrial inspection.

    \item \textbf{Language-guided two-hop grounding.}
    We introduce a compact Text$\rightarrow$Slots$\rightarrow$Pixels grounding head that links anomaly-related sentences to evidence-like latent slots and refines coarse spatial support into dense anomaly masks via channel--spatial modulation and lightweight decoding.

    \item \textbf{Executable-Rule GRPO for protocol-aligned and self-consistent reporting.}
    We design verifiable reward rules to enforce structured-format validity, grounding alignment, and reasoning--conclusion consistency, improving reliability and auditability beyond SFT without extra human supervision.

    \item \textbf{Extensive evaluation on industrial benchmarks.}
    We conduct comprehensive experiments across multiple industrial anomaly benchmarks and report strong zero-shot detection/localization performance together with improved report quality and faithfulness.
\end{itemize}

\section{Related Work}

\subsubsection{Multimodal Industrial Anomaly Detection (IAD)}
Industrial anomaly detection (IAD) is often studied under unsupervised/one-class settings, where models learn normality and detect defects via representation learning or reconstruction/localization cues.
MVTec AD~\cite{Bergmann_2019_CVPR} has driven substantial progress, including self-supervised representations~\cite{Li_2021_CVPR} and feature-based distribution/memory modeling for localization~\cite{Roth_2022_CVPR}.
Beyond single-view RGB inspection, recent works have explored RGB+3D / multi-sensor IAD with explicit cross-modal fusion/mapping strategies~\cite{Wang_2023_CVPR}, and benchmarks such as Real-IAD D3 and MulSen-AD further accelerate multimodal IAD research~\cite{Zhu_2025_CVPR,Li_2025_CVPR}.
However, most existing pipelines still output only anomaly scores or heatmaps, providing limited \emph{report-level} explanations grounded to defect regions and cross-modal evidence.
ZSG-IAD bridges this gap by coupling multimodal fusion with structured reporting and pixel-level grounding.

\begin{figure*}[htbp]
\centerline{\includegraphics[width=\linewidth]{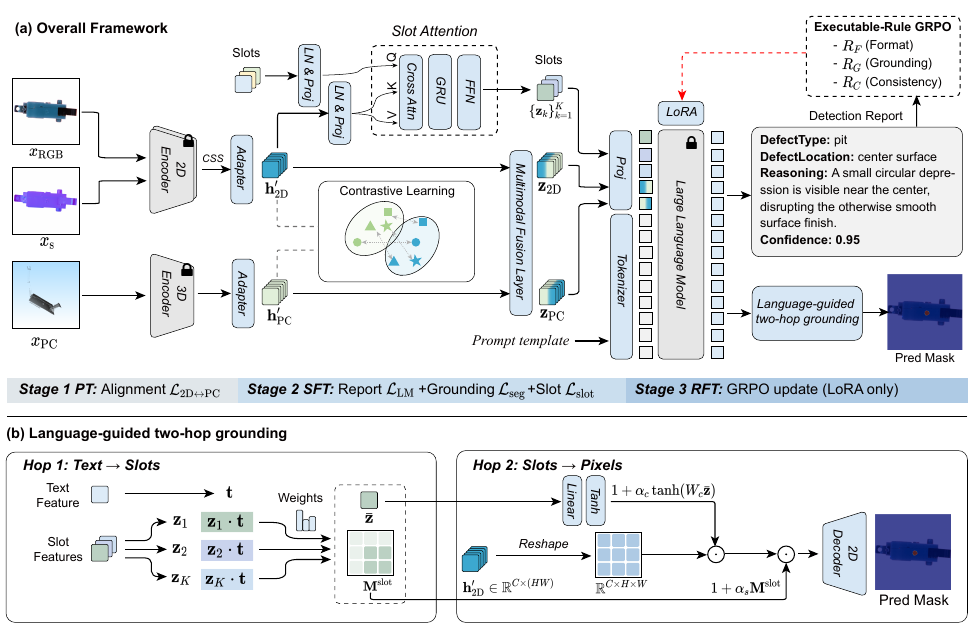}}
\caption{Overview of ZSG-IAD: frozen 2D/3D encoders with lightweight adapters, multimodal fusion, slot-based evidence tokens for structured reporting, and language-guided two-hop grounding with Executable-Rule GRPO.}
\label{fig:overview}
\end{figure*}

\subsubsection{Grounded Industrial Reporting}
Trustworthy industrial inspection increasingly requires explainability and auditability in addition to detection accuracy.
Recent surveys highlight the role of explainable AI in smart manufacturing and industrial fault diagnosis~\cite{Nikiforidis_2025_ICTExpress,Cacao_2025_JIII}, while semantic/knowledge-based representations and reasoning have been explored to support machine-checkable decision workflows in Industry 4.0 settings~\cite{Giustozzi_2024_SemanticWeb}.
Meanwhile, the broader vision-language literature has developed powerful mechanisms to connect text to visual evidence, including grounded language-image pretraining for phrase grounding/detection~\cite{Li_2022_CVPR}, text-driven dense prediction/segmentation with CLIP-style supervision~\cite{Luddecke_2022_CVPR,Rao_2022_CVPR}, and referring image segmentation with explicit language-conditioned region prediction~\cite{Yang_2022_CVPR}.
These approaches demonstrate strong general grounding capability, but they are not designed for industrial anomaly reporting where (i) targets are \emph{non-semantic} defect regions, and (ii) outputs must follow protocol/format constraints for QA workflows.
ZSG-IAD addresses this gap with a language-guided two-hop grounding design:
anomaly sentences first select evidence-like latent slots and a coarse spatial support map, and then refine it into a pixel-level anomaly mask, yielding grounded, region-specific reports that are directly auditable.

\subsubsection{Zero-Shot Learning in Industrial AI}
Industrial systems often operate in open-world environments where unseen defects emerge.
Accordingly, open-set recognition has been widely studied to handle unknown categories, from early OpenMax-style recalibration for deep classifiers~\cite{Bendale_2016_CVPR} to more recent comprehensive surveys~\cite{Geng_2021_TPAMI}.
In parallel, vision-language pretraining (e.g., CLIP) enables open-vocabulary recognition by aligning images and text in a shared embedding space~\cite{Radford_2021_ICML}.
Built on such foundations, recent top-venue methods develop zero-/few-shot anomaly classification and segmentation with prompt ensembles and patch/window-level matching, e.g., WinCLIP~\cite{Jeong_2023_CVPR}, and further improve anomaly sensitivity by anomaly-aware adaptation of CLIP~\cite{Ma_2025_CVPR}.
Nevertheless, most zero-shot anomaly methods primarily target recognition/localization quality on 2D inputs and do not produce \emph{grounded} natural-language inspection reports that connect each defect statement to localized evidence (especially under multimodal inputs).
ZSG-IAD advances zero-shot IAD by jointly generating structured defect reports and localized regions, enabling actionable inspection under unseen defect types with multimodal evidence.

\subsubsection{Reinforcement Learning for Industrial AI}
Reinforcement learning (RL) has been explored for sequential decision-making in manufacturing and edge/cloud systems, e.g., learning dispatching rules for job-shop scheduling~\cite{Zhang_2020_NeurIPS} and RL-based online coordination for microservice migration in mobile edge computing~\cite{Wang_2021_TMC}.
For LLM-based report generation, RL-style alignment/fine-tuning is commonly used to steer model outputs toward target behaviors, e.g., instruction following via human feedback~\cite{Ouyang_2022_NeurIPS} or preference optimization alternatives~\cite{Rafailov_2023_NeurIPS}, often built on policy-gradient style updates such as PPO~\cite{Schulman_2017_PPO}.
Recent works further propose efficient policy-optimization variants such as Group Relative Policy Optimization (GRPO) for large-scale LLM RL training~\cite{Shao_2024_DeepSeekMath}.
However, these approaches typically rely on preference-based or sparse task rewards, which do not directly enforce industrial protocol compliance or self-consistency between textual explanations and localization behavior.
ZSG-IAD adopts \textbf{Executable-Rule GRPO} under an RL-with-Verifiable-Rewards (RLVR) paradigm, using machine-checkable rewards to enforce (i) structured report validity, (ii) grounding alignment between the coarse language-conditioned map and the final anomaly mask, and (iii) reasoning-conclusion consistency, improving reliability and auditability without additional human preference supervision.

\section{Methodology}

\subsection{Overall Framework}
We propose \textbf{ZSG-IAD}, an end-to-end vision-language framework designed for generating \textit{interpretable, region-aware} industrial anomaly detection reports. The model takes multimodal inputs, represented as:
\[
\mathcal{X} = \{ x_{\text{RGB}}, x_s, x_{\text{PC}} \},
\]
\noindent
where \( x_{\text{RGB}} \in \mathbb{R}^{C \times H \times W} \) denotes 2D RGB images from visual inspections, \( x_s \in \mathbb{R}^{C \times H \times W} \) represents sensor data (e.g., infrared or pseudo-3D images), and \( x_{\text{PC}} \in \mathbb{R}^{N \times 3} \) is the 3D point cloud data. The framework consists of:
(1) Frozen multimodal encoders for RGB, sensor, and point cloud feature extraction;
(2) a Channel-Spatial Swapping (CSS) module to enhance the interaction between RGB and sensor features and form a unified 2D mixed feature;
(3) lightweight modality-specific adapters for feature normalization, enhancement, and alignment/projection into a common space;
(4) a multimodal fusion layer that aligns 2D and 3D features for richer context;
(5) a language model for generating structured anomaly detection reports;
(6) a language-guided two-hop detection-to-region grounding head, which converts textual descriptions into pixel-level anomaly masks for 2D images.
The overall architecture is illustrated in Fig.~\ref{fig:overview}(a).

\subsubsection{Multimodal Encoders with CSS and Adapters}
\paragraph{Frozen encoders.}
We employ a multimodal encoder architecture for processing RGB, sensor, and point cloud data.
The shared 2D encoder $E_{\text{2D}}(\cdot)$ extracts patch tokens from both $x_{\text{RGB}}$ and $x_s$,
yielding $\mathbf{h}_{\text{RGB}}$ and $\mathbf{h}_\text{s}$, respectively.
The point cloud encoder $E_{\text{3D}}(\cdot)$ produces $\mathbf{h}_{\text{PC}}$.

\paragraph{2D cross-source interaction (CSS).}
To enhance the interaction between RGB and sensor features, we adopt a Channel-Spatial Swapping (CSS) module following~\cite{Zhu_2025_CVPR}, which consists of two stages: (1) Channel Exchange and (2) Spatial Exchange.
In the Channel Exchange stage, a fraction \( p \) of the channels from the RGB feature map \( \mathbf{h}_{\text{RGB}} \) and the sensor feature map \( \mathbf{h}_{\text{s}} \) are exchanged:
\begin{equation}
    \mathbf{h}_{\text{RGB}}^{\text{mix}},\, \mathbf{h}_{\text{s}}^{\text{mix}} = \text{ChannelExchange}(\mathbf{h}_{\text{RGB}},\, \mathbf{h}_{\text{s}}).
\end{equation}
In the Spatial Exchange stage, a fraction \( p \) of the spatial features (rows or columns) are exchanged between the two feature maps based on a spatial mask:
\begin{equation}
    \mathbf{h}_{\text{2D}} = \text{SpatialExchange}(\mathbf{h}_{\text{RGB}}^{\text{mix}},\, \mathbf{h}_{\text{s}}^{\text{mix}}).
\end{equation}
Finally, the resulting features are fused to create a unified 2D mixed feature representation \(\mathbf{h}_{\text{2D}}\), enabling deeper interaction between RGB and sensor features.

\paragraph{Adapters for alignment/projection.}
After obtaining the unified 2D feature $\mathbf{h}_{\text{2D}}$, we process it through a 2D adapter module to produce the adapted representation $\mathbf{h}'_{\text{2D}}$.
In parallel, the point cloud feature $\mathbf{h}_{\text{PC}}$ is processed through a 3D adapter to produce $\mathbf{h}'_{\text{PC}}$:
\begin{equation}
    \mathbf{h}'_{\text{2D}} = \text{Adapter}_{\text{2D}}(\mathbf{h}_{\text{2D}}),\quad
    \mathbf{h}'_{\text{PC}} = \text{Adapter}_{\text{3D}}(\mathbf{h}_{\text{PC}}).
\end{equation}
Each adapter module consists of a normalization layer followed by a Multi-Layer Perceptron (MLP) and a final linear head to produce modality-specific representations.
The adapter ensures that features are normalized, enhanced via MLP, and projected into the desired output dimension, enabling better alignment and interaction between the multimodal data streams.

\subsubsection{Slot Attention for Latent Evidence Tokens}
To obtain a compact set of region-aware latent tokens, we employ Slot Attention~\cite{locatello2006object} to parse the 2D feature map $\mathbf{h}'_{\text{2D}}\in\mathbb{R}^{C\times H\times W}$ into $K$ slot representations.
We first flatten $\mathbf{h}'_{\text{2D}}$ into $\mathbf{X}\in\mathbb{R}^{(HW)\times C}$ and apply LayerNorm.
The slot matrix $\mathbf{Z}\in\mathbb{R}^{K\times D_{\text{slot}}}$ is initialized by sampling from a Gaussian distribution with learnable parameters $\mu$ and $\sigma$ (shared across slots).
Then, for $T$ refinement iterations, each slot competes to explain the input tokens via cross-attention: we project $\mathbf{X}$ to keys/values and $\mathbf{Z}$ to queries, and compute attention weights that are normalized across the $K$ slots for each spatial token.
Using these weights, we aggregate token features into slot updates by a weighted mean, and update each slot with a GRU followed by an MLP residual.
The resulting slot matrix $\mathbf{Z}$ summarizes distinct latent regions (we denote its $k$-th row as $\mathbf{z}_k$), and the attention matrix can be reshaped into $\mathbf{A}_{\text{map}}\in\mathbb{R}^{K\times H\times W}$ to visualize the spatial support of each slot.
We denote the $k$-th map in $\mathbf{A}_{\text{map}}$ as $\mathbf{A}_k\in\mathbb{R}^{H\times W}$.
Algorithm~\ref{alg:slot-attn} details the procedure.

\subsubsection{Multimodal Fusion}
After applying CSS for 2D feature fusion, we project the modality-specific features into the same embedding size by modality-specific adapters:
$\mathbf{h}'_{\text{2D}}, \mathbf{h}'_{\text{PC}} \in \mathbb{R}^{d}$.
To enable comprehensive interaction between the 2D image, and 3D point cloud modalities, we introduce a Multimodal Fusion Layer (MFL) comprising a Bidirectional Cross-Attention (BCA) mechanism. The projected features $\mathbf{h}'_{\text{2D}}$, and $\mathbf{h}'_{\text{PC}}$ are processed by the BCA mechanism, where each modality alternately serves as Query and Key/Value to compute cross-attention:
\begin{equation}
    \begin{aligned}
        & \mathbf{A}_{\text{2D} \rightarrow \text{PC}} = \text{Attn}(\mathbf{h}'_{\text{2D}}, \mathbf{h}'_{\text{PC}}, \mathbf{h}'_{\text{PC}}), \\
        & \mathbf{A}_{\text{PC} \rightarrow \text{2D}} = \text{Attn}(\mathbf{h}'_{\text{PC}}, \mathbf{h}'_{\text{2D}}, \mathbf{h}'_{\text{2D}}).
    \end{aligned}
\end{equation}

This bidirectional attention mechanism captures complex interactions between industrial data modalities, allowing visual features to inform sensor and point cloud data, and vice versa. The attention outputs are combined with residual connections to preserve modality-specific information:
\begin{equation}
    \begin{aligned}
        & \mathbf{z}_{\text{2D}} = \mathbf{h}'_{\text{2D}} + \mathbf{A}_{\text{2D} \rightarrow \text{PC}}, \\
        & \mathbf{z}_{\text{PC}} = \mathbf{h}'_{\text{PC}} + \mathbf{A}_{\text{PC} \rightarrow \text{2D}}.
    \end{aligned}
\end{equation}

\subsubsection{Textual Report Generation}
A causal language model $D(\cdot)$ generates a structured anomaly detection report $\hat{\mathcal{R}} = \{\hat{s}_1, \dots, \hat{s}_N\}$ conditioned on $\mathbf{z}_{\text{2D}}$, slot features $\{\mathbf{z}_k\}_{k=1}^{K}$, $\mathbf{z}_{\text{PC}}$.
The final multimodal features $\mathbf{z}_{\text{2D}}$, $\{\mathbf{z}_k\}_{k=1}^{K}$, and $\mathbf{z}_{\text{PC}}$ replace placeholders \texttt{<image>}, \texttt{<slot>}, and \texttt{<point cloud>} in the input prompt templates. An example prompt for anomaly detection is:
\begin{quote}
\small
``Given the visual inspection \texttt{<image>}, relevant features \texttt{<slot>}, and point cloud data \texttt{<point cloud>}, what is the most probable anomaly and its location?''
\end{quote}

\subsection{Language-guided Two-hop Grounding}
\label{sec:der}
Given the text feature $\mathbf{t}$ from the report decoder, we design a language-guided two-hop grounding module (Fig.~\ref{fig:overview}(b)) that maps the text feature to a pixel-level anomaly mask. As summarized in Alg.~\ref{alg:twohop}, the grounding proceeds in two hops: (1) \textbf{text-to-slot grounding}, which selects relevant latent evidence slots and yields a coarse spatial support, and (2) \textbf{slot-to-region decoding}, which refines the coarse cue into a dense anomaly mask via lightweight gating and decoding.

\begin{algorithm}[t]
\caption{Slot Attention on 2D Feature Map}
\label{alg:slot-attn}
\begin{algorithmic}[1]
\REQUIRE $\mathbf{h}'_{\text{2D}}\in\mathbb{R}^{C\times H\times W}$; number of slots $K$; iterations $T$;
learnable $\mu,\sigma\in\mathbb{R}^{D_{\text{slot}}}$.
\ENSURE slot features $\mathbf{Z}\in\mathbb{R}^{K\times D_{\text{slot}}}$; attention maps $\mathbf{A}_{\text{map}}\in\mathbb{R}^{K\times H\times W}$.

\STATE $\mathbf{X} \leftarrow \mathrm{LN}(\mathrm{Flatten}(\mathbf{h}'_{\text{2D}})) \in\mathbb{R}^{(HW)\times C}$
\STATE $\mathbf{Z} \sim \mathcal{N}(\mu,\mathrm{diag}(\sigma)) \in \mathbb{R}^{K\times D_{\text{slot}}}$

\FOR{$t=1$ to $T$}
    \STATE $\mathbf{Z}_{\text{prev}} \leftarrow \mathbf{Z}$
    \STATE $\mathbf{Z} \leftarrow \mathrm{LN}(\mathbf{Z})$
    \STATE $\mathbf{A} \leftarrow \mathrm{Softmax}\!\Big(\frac{k(\mathbf{X})\,q(\mathbf{Z})^\top}{\sqrt{D}},\ \text{axis}=\text{slots}\Big)$
    \STATE $\mathbf{U} \leftarrow \mathrm{WeightedMean}(\text{weights}=\mathbf{A}+\epsilon,\ \text{values}=v(\mathbf{X}))$
    \STATE $\mathbf{Z} \leftarrow \mathrm{GRU}(\text{state}=\mathbf{Z}_{\text{prev}},\ \text{inputs}=\mathbf{U})$
    \STATE $\mathbf{Z} \leftarrow \mathbf{Z} + \mathrm{MLP}(\mathrm{LN}(\mathbf{Z}))$
\ENDFOR

\STATE $\mathbf{A}_{\text{map}} \leftarrow \mathrm{Reshape}(\mathbf{A}^\top)\in\mathbb{R}^{K\times H\times W}$
\RETURN $\mathbf{Z},\ \mathbf{A}_{\text{map}}$
\end{algorithmic}
\end{algorithm}

\subsubsection{Inputs and representations}
From the 2D encoder and adapter, we obtain a feature map $\mathbf{h}'_{\text{2D}}\in\mathbb{R}^{C\times H\times W}$ (with $H=W=28$ in our implementation). A slot-attention style module further produces $K$ slot features $\{\mathbf{z}_k\in\mathbb{R}^{d}\}_{k=1}^{K}$ and their corresponding attention maps $\{\mathbf{A}_k\in\mathbb{R}^{H\times W}\}_{k=1}^{K}$, where each slot is expected to capture a coherent part, object, or defect-related pattern. For the report representation, we use the hidden-state mean pooled feature from the language model as the text feature $\mathbf{t}\in\mathbb{R}^{d}$.

\subsubsection{Hop-1: Text-to-slot grounding (coarse evidence selection)}
The first hop computes a soft selection over slots conditioned on the sentence. Concretely, we measure similarity between the text feature $\mathbf{t}$ and each slot feature $\mathbf{z}_k$, and normalize across slots to obtain weights $\mathbf{w}\in\mathbb{R}^{K}$ (Alg.~\ref{alg:twohop}, line 1). These weights are used to form (i) a slot-aggregated feature $\bar{\mathbf{z}}$ and
(ii) a coarse spatial support map $\mathbf{M}^{\text{slot}}$ by combining the slot attention maps
(Alg.~\ref{alg:twohop}, line 2).
Intuitively, Hop-1 answers: which latent slots (and roughly where) are most relevant to the text?

\subsubsection{Hop-2: Slot-to-region decoding (fine mask refinement)}
The second hop refines the coarse cue into a dense anomaly mask.
We convert the aggregated slot feature $\bar{\mathbf{z}}$ into a channel gate $\mathbf{g}_c\in\mathbb{R}^{C}$
and the coarse map $\mathbf{M}^{\text{slot}}$ into a spatial gate $\mathbf{g}_s\in\mathbb{R}^{H\times W}$
(Alg.~\ref{alg:twohop}, lines 3--4). After broadcasting both gates, we modulate the 2D feature map and feed it into a
lightweight segmentation decoder $\phi_{\text{seg}}$ to predict low-resolution mask logits, which are then upsampled
and passed through a sigmoid to obtain the final anomaly mask $\hat{\mathbf{M}}$
(Alg.~\ref{alg:twohop}, lines 5--7).
This design keeps the grounding head compact while enabling language-conditioned, region-specific localization.

\begin{algorithm}[t]
\caption{Language-guided Two-hop Grounding}
\label{alg:twohop}
\begin{algorithmic}[1]
\REQUIRE 2D feature map $\mathbf{F}\in\mathbb{R}^{C\times H\times W}$; slot features $\{\mathbf{z}_k\}_{k=1}^{K}$;
slot attention maps $\{\mathbf{A}_k\}_{k=1}^{K}$; text feature $\mathbf{t}\in\mathbb{R}^{d}$.
\ENSURE anomaly mask $\hat{\mathbf{M}}\in[0,1]^{H_0\times W_0}$.

\STATE $\mathbf{w} \leftarrow \mathrm{Softmax}\big(\langle \mathrm{norm}(\mathbf{t}), \mathrm{norm}(\mathbf{z}_k)\rangle/\tau\big) \in\mathbb{R}^{K}$
\STATE $\bar{\mathbf{z}} \leftarrow \sum_{k=1}^{K} w_{k}\mathbf{z}_k \quad$; $\quad \mathbf{M}^{\text{slot}} \leftarrow \mathrm{Norm}\big(\sum_{k=1}^{K} w_{k}\mathbf{A}_k\big)$
\STATE $\mathbf{g}_c \leftarrow 1 + \alpha_c \tanh(W_c\bar{\mathbf{z}})\in\mathbb{R}^{C}$
\STATE $\mathbf{g}_s \leftarrow 1 + \alpha_s \mathbf{M}^{\text{slot}}\in\mathbb{R}^{H\times W}$
\STATE $\tilde{\mathbf{F}} \leftarrow \mathbf{F}\odot \mathrm{Broadcast}(\mathbf{g}_c)\odot \mathrm{Broadcast}(\mathbf{g}_s)$
\STATE $\mathbf{L} \leftarrow \phi_{\text{seg}}(\tilde{\mathbf{F}})\in\mathbb{R}^{1\times H\times W}$
\STATE $\hat{\mathbf{M}} \leftarrow \sigma\big(\mathrm{Upsample}(\mathbf{L}, H_0, W_0)\big)$
\RETURN $\hat{\mathbf{M}}$
\end{algorithmic}
\end{algorithm}

\subsection{Executable-Rule GRPO}
\label{sec:grpo}

To improve diagnostic faithfulness beyond Supervised Fine-Tuning (SFT), we further adopt reinforcement fine-tuning with Group-Relative Policy Optimization (GRPO)~\cite{Shao_2024_DeepSeekMath} under an RL-with-Verifiable-Rewards (RLVR) paradigm. During GRPO, we freeze all non-adapter weights and update only LoRA parameters,
while a frozen SFT model is used as the reference policy.

Given a multimodal query $\mathbf{Q}$ and a generated anomaly report $\mathbf{A}$,
the optimization objective is
\begin{equation}
\max_{\pi_{\theta}}
\mathbb{E}_{\mathbf{A} \sim \pi_{\theta}(\cdot \mid \mathbf{Q})}
\Big[
R(\mathbf{Q}, \mathbf{A})
-
\beta\, \mathrm{KL}
\big(
\pi_{\theta}(\cdot \mid \mathbf{Q})
\;\|\;
\pi_{\text{ref}}(\cdot \mid \mathbf{Q})
\big)
\Big],
\label{eq:grpo_obj}
\end{equation}
where $\pi_{\theta}$ denotes the current policy,
$\pi_{\text{ref}}$ is the SFT-trained reference,
$R(\cdot)$ is a verifiable reward,
and $\beta$ controls the strength of KL regularization.

Unlike actor--critic methods, GRPO evaluates multiple candidate responses
$\{o_1,\dots,o_G\}$ generated for the same query and performs
group-wise reward normalization:
\begin{equation}
\tilde{r}_i = \frac{r_i - \mu_r}{\sigma_r + \epsilon},
\end{equation}
where $\mu_r$ and $\sigma_r$ are the mean and standard deviation of rewards within the group.
This removes the need for an explicit critic and stabilizes training in
domain-specific industrial settings.

We design rewards that are fully executable and automatically verifiable,
without relying on additional human annotations.
The total reward is defined as
\begin{equation}
R = w_F R_F + w_G R_G + w_C R_C,
\end{equation}
where each component captures a complementary aspect of diagnostic faithfulness.

\subsubsection{Structured Format Reward ($R_F$)}
We enforce a minimal, machine-checkable report schema with four required fields:
\texttt{<DefectType>}, \texttt{<DefectLocation>}, \texttt{<Reasoning>}, and \texttt{<Confidence>}.
A field is considered valid if it can be parsed by a deterministic checker and satisfies
predefined constraints (e.g., non-empty strings for \texttt{<DefectType>} and \texttt{<Reasoning>},
a parsable location format for \texttt{<DefectLocation>}, and a numerical \texttt{<Confidence>} in $[0,1]$).
The reward is defined as
\[
R_F =
\begin{cases}
1, & \text{if all four fields are present and valid}, \\
0, & \text{otherwise}.
\end{cases}
\]
This encourages standardized, interpretable anomaly reports suitable for industrial auditing.

\subsubsection{Grounding Alignment Reward ($R_G$)}
To encourage spatially faithful descriptions, we introduce a grounding alignment reward that enforces consistency between language-conditioned grounding and the final predicted anomaly region.
Specifically, for each generated anomaly report, the model produces a language-guided coarse grounding map via the text-to-slot mechanism, as well as a refined anomaly mask from the slot-to-region decoder. $R_G$ is defined based on the overlap between the coarse grounding map and the final segmentation mask (or ground-truth mask when available), measured by a soft Dice or IoU score:
\begin{equation}
R_G = \mathrm{Overlap}\big(\mathbf{M}^{\text{slot}}, \hat{\mathbf{M}}\big).
\end{equation}
This reward penalizes reports whose implied spatial focus is inconsistent with the model’s own localization output, encouraging region-aware and self-consistent explanations.

\subsubsection{Logical Consistency Reward ($R_C$)}
To prevent logically inconsistent outputs,
we assess whether the predicted defect conclusion is entailed by the preceding reasoning.
We employ a pretrained Natural Language Inference (NLI) model~\cite{he2021debertav3}
to compute
\begin{equation}
R_C = \mathrm{NLI}\big(\text{Reasoning} \Rightarrow \text{DefectType}\big),
\end{equation}
which yields higher rewards when the reported defect type is supported by the stated evidence,
and lower rewards when contradictions are detected (e.g., describing clear defect cues while concluding \texttt{Normal}).

Overall, the structured format reward enforces syntactic validity,
the grounding alignment reward ensures spatial faithfulness,
and the consistency reward promotes logical coherence between explanation and conclusion.
Together with the KL regularizer to the SFT reference policy,
Executable-Rule GRPO improves the reliability and auditability of
zero-shot industrial anomaly detection without introducing additional supervision.

\subsection{Training Strategy}
\label{sec:training}

We adopt a three-stage training pipeline: Pre-Training (PT), Supervised Fine-Tuning (SFT), and Reinforcement Fine-Tuning (RFT), to progressively equip ZSG-IAD with robust 2D-3D representation alignment, grounded anomaly localization via language-guided two-hop grounding, and faithful and auditable diagnostic reasoning via executable-rule reinforcement fine-tuning.

\subsubsection{Stage 1: Pre-Training (PT)}
\label{sec:pt}
To initialize modality-specific encoders and obtain a shared 2D--3D embedding space, we perform lightweight contrastive pre-training on paired 2D inputs (RGB and sensor images) and point clouds.
Given a batch of paired samples $\{(\mathbf{x}_{\text{2D}}^{(i)},\mathbf{x}_{\text{PC}}^{(i)})\}_{i=1}^{N}$, we encode them into $\ell_2$-normalized embeddings $\mathbf{h}_{\text{2D}}^{(i)}$ and $\mathbf{h}_{\text{PC}}^{(i)}$ and compute similarity
$s_{ij} = (\mathbf{h}_{\text{2D}}^{(i)} \cdot \mathbf{h}_{\text{PC}}^{(j)})/\tau$.
We optimize a symmetric InfoNCE objective:
\begin{equation}
\mathcal{L}_{\text{PT}}
=\frac{1}{2}\left(\mathcal{L}_{\text{2D}\rightarrow \text{PC}}+\mathcal{L}_{\text{PC}\rightarrow \text{2D}}\right),
\end{equation}
where
\begin{equation}
\mathcal{L}_{\text{2D}\rightarrow \text{PC}}
=
-\frac{1}{N}
\sum_{i=1}^{N}
\sum_{j=1}^{N}
y^{\text{2D}\rightarrow \text{PC}}_{ij}
\log
\frac{\exp(s_{ij})}{\sum_{k=1}^{N}\exp(s_{ik})},
\label{eq:pt_i2t}
\end{equation}
\begin{equation}
\mathcal{L}_{\text{PC}\rightarrow \text{2D}}
=
-\frac{1}{N}
\sum_{j=1}^{N}
\sum_{i=1}^{N}
y^{\text{PC}\rightarrow \text{2D}}_{ji}
\log
\frac{\exp(s_{ij})}{\sum_{k=1}^{N}\exp(s_{kj})},
\label{eq:pt_t2i}
\end{equation}
following the standard contrastive paradigm~\cite{Radford_2021_ICML}.
For training stability and larger effective negatives, we adopt BLIP-style momentum encoders and a feature queue~\cite{li2022blip}.
This stage provides aligned 2D-3D representations for downstream training.

\subsubsection{Stage 2: Supervised Fine-Tuning (SFT)}
\label{sec:sft}
In SFT, we train the anomaly reporting and grounding components using paired multimodal inputs and anomaly reports.
We freeze the majority of backbone parameters and jointly optimize lightweight modules, including
projection layers, the cross-modal fusion block, the slot-attention module, and the two-hop grounding decoder.
The language decoder is fine-tuned using LoRA-style adapters (and light parameters such as norms/biases), while keeping the main LLM weights fixed.

Given a multimodal query $\mathbf{Q}$ and a target report $\mathbf{T}_A=\{y_s\}_{s=1}^{S}$,
we optimize the standard autoregressive negative log-likelihood:
\begin{equation}
\mathcal{L}_{\text{LM}}
=
-\mathbb{E}_{(\mathbf{Q},\mathbf{T}_A)\sim\mathcal{D}}
\sum_{s=1}^{S}
\log \pi_{\theta}\!\left(y_s \mid \mathbf{Q}, y_{<s}\right),
\label{eq:sft_lm}
\end{equation}
where $\pi_{\theta}$ denotes the report decoder.

In parallel, we supervise the language-guided two-hop grounding module (Sec.~\ref{sec:der}) with a segmentation loss.
Given the predicted anomaly mask $\hat{\mathbf{M}}\in[0,1]^{H_0\times W_0}$ and ground-truth mask $\mathbf{M}^*\in\{0,1\}^{H_0\times W_0}$,
we use a Dice+Binary-Cross-Entropy loss:
\begin{equation}
\mathcal{L}_{\text{seg}}
=
\lambda_{\text{dice}}\bigl(1-\mathrm{Dice}(\hat{\mathbf{M}},\mathbf{M}^*)\bigr)
+
\lambda_{\text{bce}}\mathrm{BCE}(\hat{\mathbf{M}},\mathbf{M}^*).
\label{eq:sft_seg}
\end{equation}
To prevent segmentation supervision from degrading language modeling representations, we detach the sentence feature used for grounding when computing $\mathcal{L}_{\text{seg}}$; thus segmentation gradients only update grounding-related modules.

We further add a lightweight slot regularization term $\mathcal{L}_{\text{slot}}$ to encourage diverse and spatially compact slot attention maps.
The total SFT objective is:
\begin{equation}
\mathcal{L}_{\text{SFT}}=\mathcal{L}_{\text{LM}}+\lambda_{\text{twohop}}\mathcal{L}_{\text{seg}}+\lambda_{\text{slot}}\mathcal{L}_{\text{slot}}.
\label{eq:sft_total}
\end{equation}

\subsubsection{Stage 3: Reinforcement Fine-Tuning (RFT)}
\label{sec:rft}
Finally, we perform reinforcement fine-tuning using the executable-rule GRPO described in Sec.~\ref{sec:grpo}.
We freeze all non-adapter weights and update only LoRA parameters, with the SFT model as the reference policy $\pi_{\text{ref}}$.
For each multimodal query $\mathbf{Q}$, we sample a group of $G$ candidate reports, score them with our executable rewards, and apply group-relative updates with a KL anchor to $\pi_{\text{ref}}$.
This stage improves protocol adherence (schema validity), grounding alignment, and reasoning-conclusion consistency beyond SFT.

In conclusion, PT aligns 2D and 3D instance embeddings with momentum-stabilized contrastive learning;
SFT teaches the model to generate grounded anomaly reports while supervising two-hop localization;
RFT further refines behavior using executable industrial rules, yielding transparent and auditable zero-shot anomaly reports without additional human preference labels.

\begin{table}[t]
\centering
\footnotesize
\setlength{\tabcolsep}{3pt}
\renewcommand{\arraystretch}{1.12}
\caption{Source composition of MM-IAD-ReportInstruct-12K and MM-IAD-ReportBench}
\begin{tabular}{p{0.48\linewidth}|c|c|c}
\hline
\textbf{Source} & \textbf{\#Sel.} & \textbf{\#Cats.} & \textbf{Mask/3D (\%)} \\
\hline
Real-IAD D$^3$~\cite{Zhu_2025_CVPR} & 6{,}000 & 15 & 40.8 / 100 \\
MVTec 3D-AD~\cite{bergmann2021mvtec}          & 3{,}000 & 10 & 22.9 / 100 \\
Eyecandies~\cite{bonfiglioli2022eyecandies}$^\dagger$ & 3{,}000 & 10 & 13.3 / 100 \\
\hline
Real-IAD D$^3$~\cite{Zhu_2025_CVPR} & 1{,}000 & 5  & 40.8 / 100 \\
MulSen-AD~\cite{Li_2025_CVPR}$^\ddagger$        & 1{,}000 & 15 & 24.3 / 100 \\
\hline
\end{tabular}

\vspace{0.5mm}
\raggedright
\scriptsize $^\dagger$ Eyecandies provides depth/normal maps; point clouds are reconstructed from depth.\\
\scriptsize $^\ddagger$ MulSen-AD masks are modality-specific (annotated only when the anomaly is visible in that modality); Mask ratio here is a coarse abnormal/total estimate.
\label{tab:mmiad_source_singlecol}
\end{table}

\begin{table*}[t]
\centering
\caption{Performance on MM-IAD-ReportBench. We report detection, localization, report quality, and faithfulness metrics. For methods that cannot produce pixel-level masks or structured reports, the corresponding metrics are marked as ``--''. * indicates the model is fine-tuned on our dataset. All metrics are reported in percentage (\%).}
\setlength{\tabcolsep}{3.5pt}
\renewcommand{\arraystretch}{1.05}
\begin{tabular}{l|cccc|ccc|ccc|cc}
\toprule
\textbf{Model} &
\textbf{Accuracy} & \textbf{Precision} & \textbf{Recall} & \textbf{F1-score} &
\textbf{P-AUROC} & \textbf{Dice} & \textbf{IoU} &
\textbf{BERTScore} & \textbf{ROUGE-L} & \textbf{Schema} &
\textbf{Align-IoU} & \textbf{NLI} \\
\midrule
LLaVA-OV-0.5B~\cite{li2024llavaonevision} & 74.3 & 76.4 & 61.7 & 68.2 & -- & -- & -- & 41.5 & 27.8 & 72.6 & -- & 78.2 \\
LLaVA-OV-7B~\cite{li2024llavaonevision} & 78.3 & 82.1 & 70.9 & 76.1 & -- & -- & -- & 47.2 & 30.5 & 78.7 & -- & 82.4 \\
Qwen2-VL-2B~\cite{wang2024qwen2vl} & 74.2 & 85.0 & 74.8 & 79.6 & -- & -- & -- & 50.4 & 33.2 & 88.5 & -- & 81.8 \\
Qwen2-VL-7B~\cite{wang2024qwen2vl} & 75.2 & 85.4 & 78.2 & 81.7 & -- & -- & -- & 52.9 & 35.8 & 89.8 & -- & 82.3 \\
InternVL-2-8B~\cite{opengvlab2024internvl2} & 73.6 & 85.7 & 73.9 & 79.3 & -- & -- & -- & 49.6 & 32.8 & 86.8 & -- & 89.0 \\
Idefics2-8B~\cite{laurencon2024matters}   & 72.4 & 82.7 & 69.8 & 75.7 & -- & -- & -- & 46.8 & 31.0 & 82.0 & -- & 87.2 \\
GPT-4o~\cite{hurst2024gpt}        & 76.3 & \textbf{87.6} & 72.1 & 79.1 & -- & -- & -- & 52.7 & \textbf{36.9} & 91.2 & -- & 89.5 \\
\midrule
PatchCore~\cite{Roth_2022_CVPR} & 79.5 & 83.8 & 80.2 & 82.0 & 96.3 & 71.2 & 55.3 & -- & -- & -- & -- & -- \\
FastFlow~\cite{Yu_2021_FastFlow}  & 78.0 & 82.0 & 78.5 & 80.2 & 95.2 & 68.5 & 52.0 & -- & -- & -- & -- & -- \\
EfficientAD~\cite{Batzner_2024_EfficientAD} & 80.2 & 84.2 & 81.0 & 82.6 & \underline{96.9} & 72.0 & 56.5 & -- & -- & -- & -- & -- \\
\midrule
Qwen2-VL-7B* & 77.8 & 84.2 & 81.8 & 83.0 & -- & -- & -- & 56.8 & 36.2 & 95.6 & -- & 87.5 \\
\textbf{ZSG-IAD (RGB-only)} & 78.5 & 82.5 & 84.1 & 83.3 & 93.7 & 73.2 & 57.7 & 57.5 & 34.5 & 97.8 & 81.4 & 89.6 \\
\textbf{ZSG-IAD (RGB+S)} & 80.3 & 84.0 & 85.4 & 84.7 & 95.2 & \underline{76.0} & \underline{61.3} & 58.8 & 35.4 & 98.1 & \underline{86.5} & 91.3 \\
\textbf{ZSG-IAD (RGB+PC)} & \underline{81.2} & 85.1 & \underline{86.2} & \underline{85.6} & 96.4 & 75.6 & 60.8 & \underline{59.5} & \underline{36.7} & \underline{99.3} & 86.4 & \underline{92.5} \\
\textbf{ZSG-IAD (RGB+S+PC)} & \textbf{82.4} & \underline{86.7} & \textbf{88.9} & \textbf{87.8} & \textbf{97.6} & \textbf{77.6} & \textbf{63.4} & \textbf{60.1} & 36.6 & \textbf{99.6} & \textbf{87.2} & \textbf{93.8} \\
\bottomrule
\end{tabular}
\label{tab:bench_main}
\end{table*}

\begin{figure*}
    \centering
    \includegraphics[width=\linewidth]{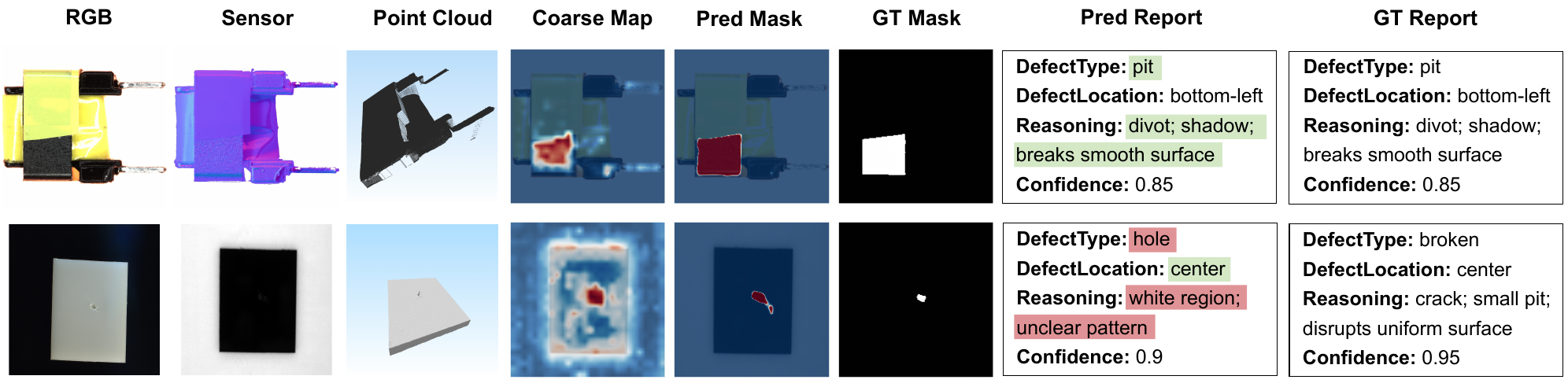}
    \caption{Qualitative examples of multimodal grounded reporting. Each example is arranged from left to right as: RGB, sensor, point cloud, coarse map, predicted mask, ground-truth mask, predicted report, and ground-truth report. We display key evidence from Reasoning for readability.}
    \label{fig:cases}
\end{figure*}

\begin{table}[htbp]
\centering
\caption{Quantitative comparison of Image-level AUROC on different ZSAD methods. The best and the second-best results are bolded and underlined, respectively. }
\begin{tabular}{l|c|c|c|c|c}
\toprule
\textbf{Model} & \textbf{VisA} & \textbf{AITEX} & \textbf{ELPV} & \textbf{BTAD} & \textbf{MPDD} \\
\midrule
CLIP & 66.4 & 71.0 & 59.2 & 34.5 & 54.3 \\
CoOp & 62.8 & 66.2 & 73.0 & 66.8 & 55.1 \\
WinCLIP & 78.8 & \textbf{73.0} & 74.0 & 68.2 & 63.6 \\
APRIL-GAN & 78.0 & 57.6 & 65.5 & 73.6 & 73.0 \\
AnoVL & 79.2 & 72.5 & 70.6 & 80.3 & 68.9 \\
AnomalyCLIP & 82.1 & 62.2 & \underline{81.5} & 88.3 & \underline{77.0} \\
AdaCLIP & \underline{85.8} & 64.5 & 79.7 & \underline{88.6} & 76.0 \\
\textbf{ZSG-IAD} & \textbf{90.4} & \underline{71.1} & \textbf{85.1} & \textbf{88.9} & \textbf{82.3} \\
\bottomrule
\end{tabular}
\label{tab:zsad_comparison}
\end{table}

\begin{figure}[ht]
    \centering
    \includegraphics[width=\linewidth]{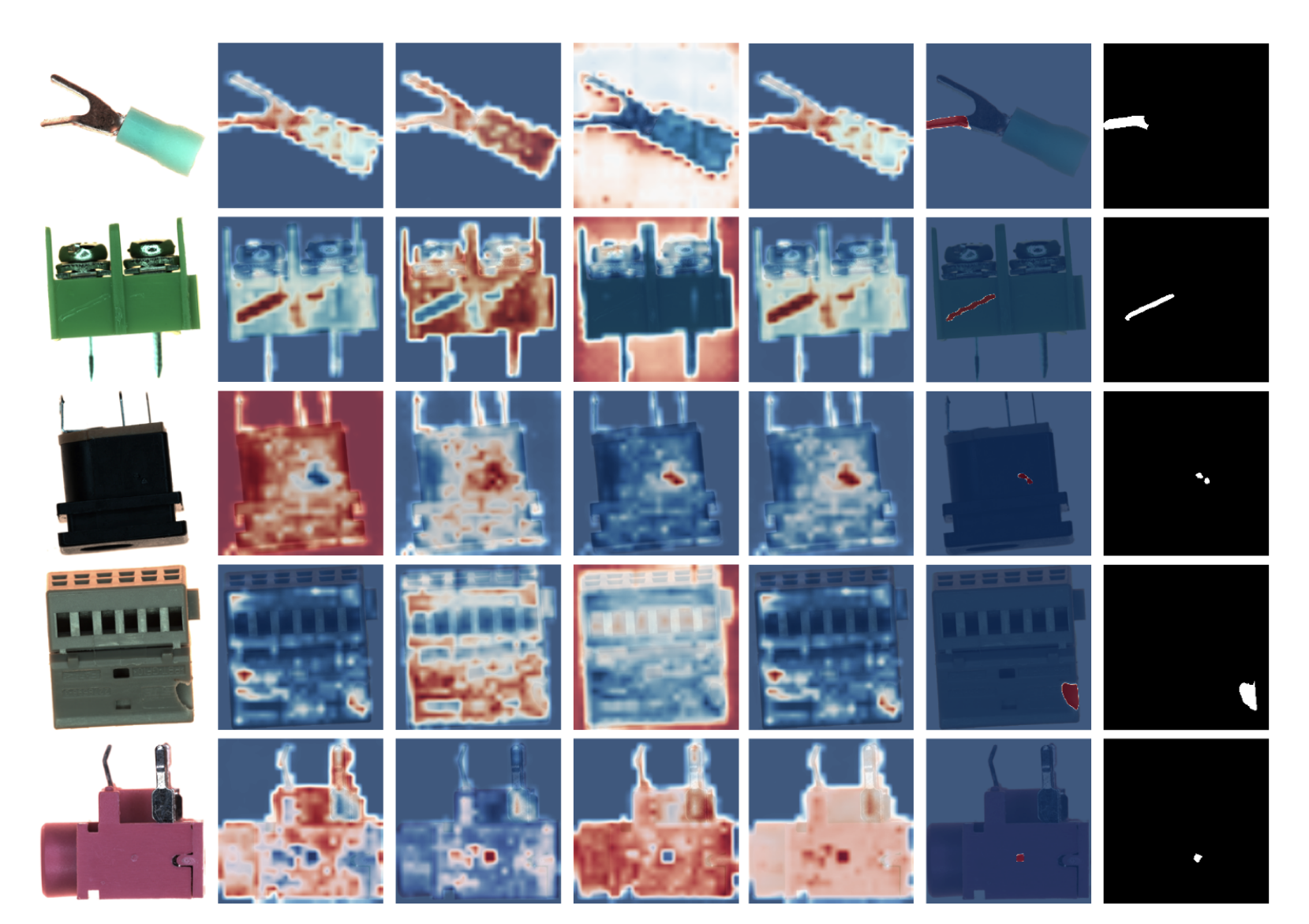}
    \caption{Qualitative visualization of language-guided two-hop grounding.
    Each example is arranged from left to right as: \textbf{(1)} RGB input; \textbf{(2--4)} the top-3 Slot Attention maps; \textbf{(5)} the aggregated coarse support map $\mathbf{M}^{\text{slot}}$; \textbf{(6)} the final pixel-level prediction $\hat{\mathbf{M}}$; and \textbf{(7)} the ground-truth mask.}
    \label{fig:slot}
\end{figure}

\section{Experiments}
\label{sec:exp}
We evaluate ZSG-IAD toward our core goal: multimodal grounded reporting with actionable localization.
The experiments are organized as follows:
(i) end-to-end detection, localization, and structured reporting on MM-IAD-ReportBench;
(ii) analysis of grounding quality and component contributions;
(iii) the effect of Executable-Rule GRPO;
(iv) robustness to missing modalities; and
(v) zero-shot transfer on standard ZSAD benchmarks as external generalization evidence.

Specifically, we answer:
(RQ1) Can ZSG-IAD jointly achieve detection, localization, and structured reporting on MM-IAD-ReportBench?
(RQ2) How do multimodal fusion, slot tokens, and two-hop grounding affect localization and faithfulness?
(RQ3) Does Executable-Rule GRPO improve schema validity and reasoning/grounding consistency beyond SFT?
(RQ4) How robust is ZSG-IAD to missing modalities at inference time?
(RQ5) Does ZSG-IAD maintain strong zero-shot performance on standard ZSAD benchmarks?

\subsection{Datasets and Benchmarks}
\label{sec:dataset}

\subsubsection{MM-IAD-ReportInstruct-12K}
We build MM-IAD-ReportInstruct-12K, a multimodal instruction-following dataset for industrial anomaly understanding and grounded reporting.
It integrates samples from Real-IAD D$^3$~\cite{Zhu_2025_CVPR}, MVTec-3D AD~\cite{bergmann2021mvtec}, Eyecandies~\cite{bonfiglioli2022eyecandies}, covering diverse industrial domains (e.g., surface defects, manufacturing parts, sensor-based inspection scenarios) and multiple anomaly patterns (e.g., scratches, dents, contamination, missing parts, structural irregularities).
Each instance contains multimodal observations, including RGB inspection images, sensor-derived images (e.g., infrared or pseudo-3D), and aligned 3D point clouds.
All modalities are preprocessed to ensure consistent resolutions and alignment, and the final samples are packaged as multimodal ``queries'' that mimic real inspection inputs.
To enable machine-checkable reporting, each instance is annotated with a structured report schema consisting of four fields:
\texttt{<DefectType>}, \texttt{<DefectLocation>}, \texttt{<Reasoning>}, and \texttt{<Confidence>}.
\texttt{<DefectType>} specifies the defect category (or \texttt{None}), \texttt{<DefectLocation>} provides a region-level description (e.g., ``upper-left edge'', ``center surface''), \texttt{<Reasoning>} summarizes evidence cues (appearance, sensor response, geometric abnormality), and \texttt{<Confidence>} provides a calibrated confidence estimate.
We build the four-field reports using a semi-automatic reporter–validator pipeline: defect types are mapped from source annotations, locations are derived from segmentation masks when available, and executable validators enforce protocol compliance; unresolved cases are manually audited.
For samples with segmentation annotations, we additionally include pixel-level anomaly masks to supervise and evaluate grounded localization.
MM-IAD-ReportInstruct-12K is used for supervised fine-tuning of structured report generation and for training the language-guided two-hop grounding head.

\subsubsection{MM-IAD-ReportBench}
We construct MM-IAD-ReportBench as a unified benchmark to evaluate multimodal anomaly detection, localization, and structured grounded reporting under a standardized protocol.
It contains 2,000 samples from Real-IAD D$^3$~\cite{Zhu_2025_CVPR}, and MulSen AD~\cite{Li_2025_CVPR}, with multimodal inputs and standardized prompts, spanning multiple object categories and defect types.

Although MM-IAD-ReportInstruct-12K and MM-IAD-ReportBench are constructed from partially shared public sources (e.g., Real-IAD D$^3$),
we strictly prevent train-test contamination, as shown in Tab~\ref{tab:mmiad_source_singlecol}.
Specifically, we split the shared sources by disjoint object categories (no category overlap between training and benchmark),
and we additionally ensure no instance-level overlap between MM-IAD-ReportInstruct-12K and MM-IAD-ReportBench.
MM-IAD-ReportBench is excluded from all optimization stages and used only for evaluation.

\subsection{Evaluation Protocol and Metrics}
\label{sec:exp_metrics}
We evaluate all methods on MM-IAD-ReportBench using unified prompt templates.
For fair comparison, methods without point-cloud support are evaluated in the 2D-only setting (RGB+S), while we additionally report ZSG-IAD under multiple input modalities to quantify robustness and modality contributions.
For detection, we report image-level classification metrics (Accuracy, Precision, Recall, and F1-score).
For standard zero-shot anomaly detection (ZSAD) benchmarks (VisA, AITEX, ELPV, BTAD, and MPDD), we follow prior work and report image-level AUROC (I-AUROC).
For localization, we evaluate samples with pixel-level annotations using pixel-level AUROC (P-AUROC) computed over per-pixel anomaly scores, and Dice/IoU.
For structured reporting, we measure text quality with ROUGE-L and BERTScore on the concatenated four-field report.
Finally, we assess faithfulness with three metrics: Schema Valid Rate (percentage of outputs parsable into all required fields), Align-IoU (IoU between the Hop-1 coarse support map $\mathbf{M}^{slot}$ and the final mask $\hat{\mathbf{M}}$), and NLI (whether \texttt{<DefectType>} is entailed by \texttt{<Reasoning>}).

\subsection{Implementation Details}
\label{sec:exp_impl}

All experiments are conducted on NVIDIA RTX 3090 GPUs.
We use LLaMA 3.2~\cite{grattafiori2024llama3herdmodels} as the report decoder with rank-8 LoRA~\cite{hu2022lora}.
The 2D encoder is ViT-based for RGB and sensor images, and the 3D encoder is PointTransformer-based~\cite{zhao2021point}.
We set the embedding dimension to 768 and the contrastive temperature $\tau=0.07$.
Executable-Rule GRPO uses group size $G=4$, clipping $\epsilon=0.2$, and KL coefficient $\beta=0.1$.
All 2D RGB images and sensor data are preprocessed using standard normalization, and point cloud data is preprocessed using voxelization and downsampling.

\subsection{Experimental Results}

\subsubsection{Detection, Localization, and Reporting}
Table~\ref{tab:bench_main} shows that ZSG-IAD achieves the best \emph{joint} performance on MM-IAD-ReportBench when using all modalities (RGB+S+PC).
Compared with prompting generalist MLLMs, ZSG-IAD provides a better precision--recall balance (especially higher recall), which is more desirable for industrial inspection where missed defects are costly.
More importantly, ZSG-IAD produces pixel-level masks and structured reports with high schema validity and stronger reasoning--grounding consistency, enabling auditable, region-specific decisions rather than ungrounded textual claims.
Beyond generalist MLLMs, we further compare with strong anomaly detectors (PatchCore~\cite{Roth_2022_CVPR}, FastFlow~\cite{Yu_2021_FastFlow}, EfficientAD~\cite{Batzner_2024_EfficientAD}) that are trained on normal data and output dense anomaly maps.
While these baselines provide competitive detection/localization, they cannot produce protocol-compliant structured reports or guarantee reasoning--grounding faithfulness.
In contrast, ZSG-IAD jointly delivers accurate detection/localization and grounded, machine-checkable reports with higher self-consistency, enabling auditable region-specific decisions.

\subsubsection{Missing-Modality Robustness}
Table~\ref{tab:bench_main} indicates graceful degradation under missing modalities.
Point clouds contribute more to localization faithfulness than sensor images, suggesting that geometric cues help disambiguate texture-driven false alarms and better constrain defect extent.
Sensor inputs provide complementary gains mainly by enriching appearance cues, while the model remains functional under RGB-only inputs, implying that the protocol-aligned decoder and grounding head generalize beyond a single sensing channel.

\subsubsection{Zero-Shot Anomaly Detection}
\label{sec:exp_zsad}
As shown in Table~\ref{tab:zsad_comparison}, ZSG-IAD performs competitively across standard ZSAD benchmarks and ranks top on most datasets.
This suggests that multimodal alignment and grounded-report supervision do not sacrifice anomaly sensitivity; instead, they improve robustness to diverse industrial textures and unseen defect patterns.

Beyond standard ZSAD benchmarks, Tab.~\ref{tab:strict_zs_tiny} reports a strict cross-dataset zero-shot setting where we train on Real-IAD D$^3$ only and evaluate on the MulSen-AD subset, which differs in sensing conditions and annotation visibility.
ZSG-IAD maintains a clear advantage in detection (F1) while simultaneously producing actionable localization (Dice), whereas generalist MLLMs cannot output pixel masks and classical detectors cannot generate reports.
More importantly, ZSG-IAD preserves high protocol compliance (Schema) and reasoning--type consistency (NLI) under dataset shift, suggesting that the executable schema and grounded reasoning are not overfitted to a single source but transfer as verifiable decision evidence.

\begin{table}[t]
\centering
\caption{Strict cross-dataset zero-shot}
\label{tab:strict_zs_tiny}
\setlength{\tabcolsep}{4pt}
\renewcommand{\arraystretch}{1.08}
\begin{tabular}{l|c|c|c|c}
\toprule
\textbf{Model} & \textbf{F1} & \textbf{Dice} & \textbf{Schema} & \textbf{NLI} \\
\midrule
Qwen2-VL-7B (prompt) & 80.0 & -- & 89.0 & 82.8 \\
EfficientAD & 81.3 & 70.3 & -- & -- \\
\textbf{ZSG-IAD (RGB+S+PC)} & \textbf{86.2} & \textbf{75.5} & \textbf{99.0} & \textbf{92.2} \\
\bottomrule
\end{tabular}
\end{table}

\subsubsection{Ablation Studies}
\label{sec:exp_ablation}
Table~\ref{tab:ablation} reveals that slot tokens are the key bottleneck for grounding: removing slots causes the largest drop in localization and alignment, supporting their role as evidence-like latent representations bridging text and spatial cues.
Two-hop grounding consistently outperforms one-hop grounding, validating the coarse-to-fine design for both accuracy and interpretability.
Bidirectional cross-attention further improves overall stability by enabling explicit 2D-3D interaction.
Finally, the ablations on Executable-Rule GRPO confirm that each reward component contributes to structured reporting: removing the format reward $R_F$ mainly reduces schema validity, removing the grounding reward $R_G$ substantially harms grounding alignment (Align-IoU) and localization faithfulness, while removing the consistency reward $R_C$ notably degrades logical self-consistency (NLI).
Overall, GRPO regularizes protocol compliance, grounding alignment, and reasoning coherence beyond SFT.

\begin{table}[t]
\centering
\caption{Ablations on MM-IAD-ReportBench (\%).}
\setlength{\tabcolsep}{4pt}
\renewcommand{\arraystretch}{1.05}

\begin{tabular}{l|cccc}
\toprule
\textbf{Model} & \textbf{Accuracy} & \textbf{F1-score} & \textbf{P-AUROC} & \textbf{Dice} \\
\midrule
Full (RFT)        & 82.4 & 87.8 & 97.6 & 77.6 \\
w/o BCA           & 82.2 & 86.3 & 95.0 & 75.5 \\
w/o Slots         & 80.6 & 79.2 & 78.3 & 52.3 \\
One-hop grounding & 81.3 & 81.3 & 94.0 & 70.1 \\
SFT only          & 81.6 & 83.0 & 96.5 & 75.6 \\
\midrule
RFT w/o $R_F$ (format)    & 82.3 & 87.6 & 97.5 & 77.4 \\
RFT w/o $R_G$ (grounding) & 82.0 & 86.8 & 96.8 & 75.9 \\
RFT w/o $R_C$ (consist.)  & 82.2 & 87.0 & 97.3 & 77.0 \\
\bottomrule
\end{tabular}

\vspace{3pt}

\begin{tabular}{l|ccc}
\toprule
\textbf{Model} & \textbf{Schema} & \textbf{Align-IoU} & \textbf{NLI} \\
\midrule
Full (RFT)        & 99.6 & 87.2 & 93.8 \\
w/o BCA           & 99.3 & 82.6 & 90.4 \\
w/o Slots         & 97.4 & 75.0 & 88.3 \\
One-hop grounding & 98.7 & 79.5 & 90.2 \\
SFT only          & 96.2 & 83.3 & 89.5 \\
\midrule
RFT w/o $R_F$ (format)    & 97.0 & 86.5 & 93.0 \\
RFT w/o $R_G$ (grounding) & 99.4 & 83.5 & 93.1 \\
RFT w/o $R_C$ (consist.)  & 99.5 & 86.4 & 90.2 \\
\bottomrule
\end{tabular}

\label{tab:ablation}
\end{table}

\subsubsection{Qualitative Results}
\label{sec:exp_qual}

Fig.~\ref{fig:slot} visualizes the proposed Text$\rightarrow$Slots$\rightarrow$Pixels grounding.
The top-ranked slot attention maps act as candidate evidence regions; Hop-1 aggregates them into a coarse support map $\mathbf{M}^{\text{slot}}$ that reveals \emph{where} the model bases its claim, while Hop-2 refines this cue into a sharper pixel-level mask $\hat{\mathbf{M}}$ by slot-conditioned modulation, improving spatial focus and boundary quality.
Fig.~\ref{fig:cases} further illustrates multimodal grounded reporting with a representative \emph{good} case (top) and \emph{bad} case (bottom).
In the good case, cues across RGB/sensor/point cloud are consistent, leading to a compact coarse map and a mask aligned with the defect, and the structured report is supported by localized evidence.
In the bad case, the defect is small/ambiguous and competes with salient non-defect structures (e.g., edges/highlights/background patterns), causing Hop-1 to select suboptimal evidence slots and produce a biased or diffuse coarse map; this error is propagated by Hop-2, resulting in imperfect localization and a less reliable defect-type conclusion with generic reasoning.

\section{Conclusion}
We proposed ZSG-IAD, a multimodal vision--language framework for zero-shot grounded industrial anomaly detection.
Given RGB images, sensor-derived images, and 3D point clouds, ZSG-IAD generates machine-checkable structured anomaly reports together with pixel-level anomaly masks, enabling transparent and actionable inspection.
Our key technical contribution is a language-guided two-hop grounding design (Text$\rightarrow$Slots$\rightarrow$Pixels) that explicitly links anomaly descriptions to evidence-like latent slots and further refines them into fine-grained defect regions.
To improve reliability, we introduce Executable-Rule GRPO under an RL-with-Verifiable-Rewards paradigm, enforcing report-format validity, grounding alignment, and reasoning--conclusion consistency without additional human preference labels.
Extensive experiments on multiple industrial benchmarks demonstrate that ZSG-IAD achieves competitive or state-of-the-art zero-shot detection performance while producing more faithful and physically grounded reports than existing methods.
In future work, we plan to extend grounding to full 3D defect localization and explore stronger industrial protocol rules for closed-loop diagnosis and rework.

\bibliographystyle{IEEEtran}
\bibliography{refs}


 




\vfill

\end{document}